**Elsa TOLONE**
**LIGM, Université Paris-Est, France & FaMAF, Universidad Nacional de Córdoba, Argentine**
elsa.tolone@univ-paris-est.fr

**Éric de LA CLERGERIE**
**ALPAGE, INRIA Paris–Rocquencourt & Université Paris 7, France**
Eric.De_La_Clergerie@inria.fr

**Benoît SAGOT**
**ALPAGE, INRIA Paris–Rocquencourt & Université Paris 7, France**
benoit.sagot@inria.fr

## ÉVALUATION DE LEXIQUES SYNTAXIQUES
## PAR LEUR INTÉGRATION DANS L'ANALYSEUR SYNTAXIQUE FRMG

**Résumé**
Dans cet article, nous évaluons divers lexiques avec l'analyseur syntaxique FRMG : le Le*fff*, *LGLex*, le lexique syntaxique construit à partir des tables du Lexique-Grammaire du français, le lexique DICOVALENCE, ainsi qu'une nouvelle version des entrées verbales du Le*fff*, obtenues par fusion avec DICOVALENCE et validation manuelle partielle. Pour cela, tous ces lexiques ont été convertis au format du Le*fff*, le format Alexina. L'évaluation a été faite sur la partie du corpus EASy utilisé lors de la première campagne d'évaluation Passage.
Mots clés : lexiques syntaxiques, analyseur syntaxique, campagne d'évaluation, fouille d'erreurs

### 1. Introduction

De nombreux analyseurs du français ont été développés ces dernières années et il importe de les évaluer afin d'améliorer leur précision et leur robustesse sur des corpus à grande échelle. Ceci est en effet de nature à améliorer l'exploitation des annotations syntaxiques produites, d'une part pour enrichir, améliorer voire créer des ressources linguistiques et d'autre part en vue d'applications concrètes comme l'extraction d'informations ou la fouille de textes.

L'objectif de ce travail est d'évaluer divers lexiques syntaxiques par le biais de l'évaluation de l'analyseur syntaxique FRMG (de La Clergerie, 2005a ; Thomasset & Éric de La Clergerie, 2005) lorsqu'il est couplé à chacun d'entre eux. Il s'agit donc d'une évaluation *orientée-tâche* (*task-based*) de ces ressources. Nous avons évalué le Le*fff* dans sa version 3.0 (Sagot, 2010), ainsi que trois ressources obtenues en remplaçant (presque) toutes les entrées verbales du Le*fff* par d'autres ressources, à savoir successivement le lexique *LGLex* (Tolone, 2011) construit à partir des tables du Lexique-Grammaire du français, le lexique DICOVALENCE (van den Eynde & Mertens, 2006), ainsi qu'une nouvelle version des entrées verbales du Le*fff* obtenues à partir de sa fusion avec DICOVALENCE et d'un travail de validation manuelle. Pour cela, tous ces lexiques ont été convertis au format du Le*fff*, le format Alexina. L'évaluation a été faite sur la partie du corpus EASy utilisé lors de la première campagne d'évaluation Passage (Hamon *et al.*, 2008).

Nous commençons par détailler ces quatre ressources lexicales, qu'il s'agisse de lexiques déjà existants (le Le*fff*, *LGLex* ou DICOVALENCE) ou de la nouvelle version du Le*fff* (section 2.2). Nous présenterons ensuite l'analyseur FRMG et la campagne d'évaluation Passage à la section 3, avant de détailler les résultats obtenus par FRMG lorsqu'on lui fait rejouer cette campagne en le couplant successivement avec les quatre lexiques décrits à la section 4. Nous montrons en particulier que pour l'instant, c'est encore la version standard du Le*fff* qui donne les meilleurs résultats. Enfin, nous discutons à la section 5 de la fouille d'erreurs réalisée sur les sorties produites afin d'identifier les entrées lexicales verbales susceptibles d'expliquer la baisse de précision obtenue par les autres ressources par rapport au Le*fff*, puis nous concluons à la section 6.

**2. Les ressources lexicales**
2.1. Le*fff*, *LGLex* et DICOVALENCE
Nous avons utilisé les données lexicales issues de trois ressources électroniques librement disponibles :
– Le*fff* (Lexique des formes fléchies du français)[1] : Ce lexique est une ressource morphologique et syntaxique à large couverture du français, qui couvre l'ensemble des catégories (Sagot, 2010). Le Le*fff*, développé dans le formalisme lexical Alexina, est orienté vers une utilisation dans des outils de traitement automatique, mais cherche à préserver une pertinence linguistique. Il est ainsi utilisé par exemple dans des analyseurs à grande échelle pour différents formalismes (LFG, LTAG, et d'autres). Des travaux récents en ont amélioré la qualité et la couverture pour certaines classes d'entrées (constructions impersonnelles, constructions pronominales, adverbes en *-ment*, verbes en *-iser* et *-ifier*), notamment par comparaison et fusion avec d'autres ressources lexicales comme DICOVALENCE et les tables du Lexique-Grammaire (Sagot & Danlos, 2007 ; Sagot & Fort, 2007 ; Danlos & Sagot, 2008 ; Sagot & Fort, 2009).
– *LGLex*[2] : Ce lexique syntaxique a été construit à partir des tables du Lexique-Grammaire du français en un format textuel et XML (Constant & Tolone, 2010), après un travail de mise en cohérence et d'explicitation des propriétés syntaxiques dans les tables du Lexique-Grammaire (Tolone, 2011). Grâce à une définition formelle ou à une interprétation dynamique de toutes les constructions, la version texte du lexique *LGLex* a ensuite été convertie au format Alexina (Tolone & Sagot, 2011). Cela a pu être fait pour l'ensemble des verbes (issus des 67 tables regroupant 13 867 entrées, dont 5 738 entrées distinctes) et des noms prédicatifs (issus des 78 tables regroupant 12 696 entrées, dont 8 531 entrées distinctes).
– DICOVALENCE[3] : Le dictionnaire de valence verbale DICOVALENCE (van den Eynde & Mertens, 2006) est une ressource informatique qui répertorie les cadres de valence de plus de 3 700 verbes simples du français, soit plus de 8 000 entrées. Le dictionnaire explicite en outre certaines restrictions sélectionnelles, certaines formes de réalisation (pronominales, phrastiques) des termes, la possibilité d'employer le cadre valenciel dans différents types de passif, etc. La particularité essentielle du dictionnaire réside dans le fait que les informations valencielles sont représentées selon les principes de « *l'Approche Pronominale* » en syntaxe (Blanche-Benveniste *et al.*, 1984). Pour chaque place de valence (appelée *paradigme*) le dictionnaire précise le paradigme de pronoms qui y est associé et qui couvre *en intention* les lexicalisations possibles. Il précise aussi les *reformulations* possibles, comme le passif.

2.2. Construction d'une nouvelle version des entrées verbales du Le*fff* par fusion avec DICOVALENCE et validation manuelle
Bien que le principe général sous-tendant les entrées lexicales du Le*fff* soit que chaque sens distinct d'un même lemme doive correspondre à une entrée distincte, ce principe n'est respecté que très partiellement dans la version actuelle de la ressource. C'est pourtant une nécessité pour améliorer la qualité du Le*fff* comme ressource descriptive, pour permettre la prise en compte d'informations telles que les restrictions de sélection pendant ou après l'analyse syntaxique, pour coupler à terme le Le*fff* avec des ressources lexicales sémantiques, et plus

---
[1] Distribution de la version 3.0 en ligne sous licence LGPL-LR à l'adresse http://gforge.inria.fr/projects/alexina/
[2] Distribution de la version 3.3 en ligne sous licence LGPL-LR à l'adresse http://infolingu.univ-mlv.fr, Données Linguistiques > Lexique-Grammaire > Téléchargement
[3] Distribution de la version 2 en ligne sous licence LGPL-LR à l'adresse http://bach.arts.kuleuven.be/dicovalence/

généralement pour envisager l'utilisation du Le*fff* en analyse sémantique.

Nous avons effectué un premier travail dans cette direction, en cherchant à intégrer DICOVALENCE au sein du Le*fff*. En effet, DICOVALENCE distingue quant à lui les différents sens d'un même lemme verbal en plusieurs entrées. Pour cela, nous avons mis en œuvre la méthodologie décrite dans (Sagot & Danlos, 2008). Ainsi, nous avons tout d'abord converti DICOVALENCE au format Alexina, améliorant pour ce faire l'outil de conversion utilisé précédemment (Danlos & Sagot, 2008). La fusion du résultat de cette conversion avec le Le*fff* a été réalisée de la même façon que dans (Danlos & Sagot, 2008), en préservant toutes les informations issues des deux ressources (exemples, etc.). La difficulté est qu'il est fréquent qu'un lemme verbal donné ait plusieurs entrées dans DICOVALENCE et plusieurs dans le Le*fff*, ce qui rend délicate la mise en correspondance de chaque entrée de l'un avec zéro, une ou plusieurs entrées de l'autre. Nous avons donc appliqué les heuristiques décrites dans (Danlos & Sagot, 2008), qui permettent la mise en correspondance de deux entrées si les inventaires de fonctions syntaxiques de base (sujet, objets direct et indirects) sont identiques, et si l'inventaire de fonctions syntaxiques obliques (locatif, délocatif, etc.) dans l'entrée du Le*fff* est inclus dans celui issu de DICOVALENCE. Pour chaque lemme, on obtient ainsi au moins autant d'entrées que dans le lexique qui en contient le moins, et au plus la somme des nombres d'entrées dans chaque lexique, lorsqu'aucune mise en correspondance n'a fonctionné.

Pour bénéficier au mieux de la bonne qualité générale des informations syntaxiques présentes dans DICOVALENCE, nous avons décidé de réaliser une validation manuelle partielle mais significative du résultat de la fusion. Ainsi, nous avons validé manuellement toutes les entrées correspondant à des lemmes telles que le nombre d'entrées dans le lexique fusionné était strictement supérieur au maximum du nombre d'entrées entre les deux ressources. Une telle situation signifie en effet qu'au moins une entrée du Le*fff* n'a pu être mise en correspondance avec une entrée de DICOVALENCE, et inversement, ce qui fait soupçonner que la fusion s'est passée de façon incorrecte en raison d'erreurs dans l'un ou l'autre des lexiques, ou en raison de différences d'analyse (un objet indirect en *de* pour une ressource pouvant être un délocatif pour une autre, par exemple). Nous avons ainsi validé, corrigé voire fusionné manuellement toutes les entrées pour 505 lemmes verbaux, produisant ainsi 986 entrées. Par ailleurs, nous avons extrait du corpus de l'*Est Républicain* une table de fréquence des formes fléchies, qui nous a permis de dresser une liste des 100 lemmes verbaux les plus fréquents du français. Nous avons validé, corrigé, fusionné et complété manuellement toutes les entrées correspondant à ces lemmes verbaux dans le lexique fusionné.

Le résultat de ce travail est un lexique morphologique, syntaxique et sémantique composé de 12 613 entrées couvrant 7 933 lemmes verbaux distincts.

**3. L'analyseur syntaxique FRMG et la campagne d'évaluation Passage**
FRMG (FRench MetaGrammar) (de La Clergerie, 2005a ; Thomasset & Éric de La Clergerie, 2005) est un analyseur syntaxique profond à large couverture pour le français. Une description grammaticale de haut niveau sous forme de méta-grammaire sert de point de départ pour la génération d'une grammaire d'arbres adjoints (TAG, *Tree Adjoining Grammar*) (Joshi *et al.*, 1975). Cette grammaire est transformée par le système DyALog (de La Clergerie, 2005b ; de La Clergerie, 2002) en un analyseur syntaxique.

L'analyseur syntaxique FRMG découlant des phases de compilation de la métagrammaire FRMG ne peut bien sûr fonctionner seul. Il s'intègre dans une chaîne complète de traitement comprenant, en amont, le lexique syntaxique Le*fff* et les nombreux modules de SXPipe (Sagot & Boullier, 2008) en charge de la segmentation, de la correction orthographique et de la détection des entités nommées.

Pour une phrase donnée, FRMG retourne l'ensemble des analyses complètes sous forme de forêt. En cas d'échec pour une analyse complète, l'analyseur retourne un ensemble d'analyses partielles couvrant au mieux la phrase. Enfin, sous la contrainte d'un temps limite (*timeout*), si l'analyseur n'a pu conclure l'ensemble des analyses, il retourne celles déjà disponibles (mode *just-in-time*). En pratique, très peu de phrases (moins de 1%) se retrouvent sans aucune analyse (complète ou partielle).

Ensuite, la forêt d'analyse peut être convertie sous forme d'une forêt de dépendances et également désambiguïsée, en utilisant un ensemble de règles heuristiques très peu lexicalisées. On obtient ainsi une unique analyse par dépendance qui peut ensuite être convertie dans le format Passage, utilisé dans le cadre des campagnes d'évaluation Passage.

On peut voir par exemple à la figure 1 la sortie au format Passage de la phrase *Depuis quelques semaines, les rapports entre les deux camps se dégradent.*

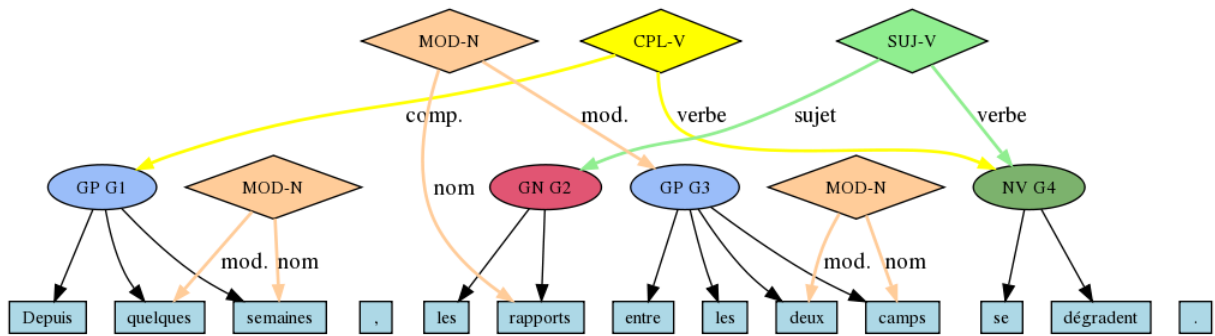

Figure 1. Exemple de sortie au format Passage

Précisons que FRMG est un logiciel libre accessible sous la GForge de l'INRIA[4]. Il est également possible de jouer avec la chaîne de traitement et de visualiser la grammaire FRMG sur http://alpage.inria.fr/frmgdemo.

La campagne d'évaluation Passage (2007-2009)[5] (Hamon *et al.*, 2008) soutenue par l'ANR a pour objectif d'évaluer les analyseurs du français, d'améliorer l'exactitude et la robustesse des analyseurs du français sur des corpus à grande échelle (100 millions de mots) ; et d'exploiter les annotations syntaxiques résultantes pour créer une ressource linguistique plus riche et plus étendue : un treebank pour le français.

Six types de constituants ont été choisis pour cette campagne : groupe nominal, (GN), noyau verbal (NV), groupe adjectival (GA), groupe adverbial (GR), groupe prépositionnel (GP) et groupe prépositionnel à noyau verbal (PV).

Les relations (dépendances entre mots pleins) à relever sont les suivantes : dépendance sujet-verbe (SUJ-V), dépendance auxiliaire-verbe (AUX-V), objet direct (COD-V), autre complément du verbe (CPL-V), modifieur du verbe (MOD-V), subordonnée (COMP), attribut du sujet ou de l'objet (ATB-SO), modifieur du nom (MOD-N), modifieur de l'adjectif (MOD-A), modifieur de l'adverbe (MOD-R), modifieur de la préposition (MOD-P), coordination (COORD), apposition (APPOS) et juxtaposition (JUXT).

Le guide d'annotation utilisé lors de la campagne Passage est le guide des annotations syntaxiques de référence PEAS[6]. Les métriques d'évaluation utilisées sont celles définies à l'occasion de la campagne EASy : la précision, le rappel et la f-mesure, avec différents modes de relâchement de contrainte sur la mesure des frontières de constituants (Paroubek *et al.*, 2005).

Pour la première campagne de Passage de 2007, la référence utilisée est un extrait d'environ un million de mots du corpus EASy (Paroubek *et al.*, 2005), de styles variés (journalistique, médical, oral, questions, littéraire, etc.), dont 4 306 phrases constituaient déjà la référence de la campagne EASy en 2004 et 400 nouvelles phrases ont été annotées manuellement depuis.

---
[4] http://gforge.inria.fr/projects/mgkit/
[5] http://atoll.inria.fr/passage/index.fr.html
[6] Disponible sur le site http://www.limsi.fr/Individu/anne/Guide/PEAS_reference_annotations_v2.2.html

Pour pouvoir tenir compte des noms prédicatifs du lexique *LGLex*, il a fallu modifier la méta-grammaire en faisant l'approximation suivante : n'importe quel déterminant peut venir de manière optionnelle devant un nom prédicatif.

**4. Résultats**

On peut voir les résultats des différents lexiques sur le corpus EASy dans le tableau 1.

| **Lexique** | **Couverture (#phrases)** | **Couverture (%)** | **Constituants (%)** | **Relations (%)** | **Temps (s)** | **Timeout (%)** |
|---|---|---|---|---|---|---|
| Le*fff* | 3 555 | 76.08 | 89.21 | 66.36 | 0.30 | 0.00 |
| Nouveau Le*fff* | 3 495 | 74.81 | 88.65 | 65.41 | 0.43 | 0.03 |
| *LGLex* | 3 437 | 73.60 | 87.97 | 63.03 | 0.84 | 0.03 |
| DICOVALENCE | 2 773 | 59.78 | 86.98 | 61.91 | 0.42 | 0.00 |

Tableau 1. Résultats comparatifs sur le corpus EASy, exprimés en terme de f-mesure

Les meilleurs résultats et performances de FRMG sont obtenus avec la version standard du Le*fff* (version 3). Ceci peut s'expliquer par le fait que les deux ressources ont été développées ensemble depuis longtemps. En comparaison, les résultats sont finalement assez proches pour les autres ressources.

Les temps d'analyse sont plus importants pour *LGLex* (avec en conséquence plus de phrases échouant pour cause de *timeout*) : ceci provient du grand nombre d'entrées par verbe dans *LGLex*. À titre de comparaison, les lemmes verbaux les plus ambigus dans le Le*fff* sont *tenir* et *(re)faire* (6 entrées), alors que dans le lexique *LGLex* il s'agit des lemmes *tenir* (53 entrées), *jouer* (44 entrées) et *prendre* (35 entrées). De plus, *LGLex* contient un grand nombre de noms prédicatifs liés à des verbes support.

DICOVALENCE a un faible taux de couverture (relativement aux autres lexiques) mais des f-mesures qui sont finalement bonnes : ceci confirme la pertinence de l'approche suivie par les développeurs de cette ressource, qui se sont concentrés sur les emplois fréquents des lemmes les plus fréquents.

Nous présentons les résultats pour quelques relations verbales dans le tableau 2.

| **Lexique** | **SUJ-V (%)** | **AUX-V (%)** | **COD-V (%)** | **CPL-V (%)** | **ATB-SO (%)** |
|---|---|---|---|---|---|
| Le*fff* | 79.29 | 91.55 | 72.48 | 62.40 | 66.46 |
| Nouveau Le*fff* | 78.76 | 91.15 | 72.18 | 62.59 | 59.45 |
| *LGLex* | 77.78 | 89.28 | 66.46 | 59.47 | 45.79 |
| DICOVALENCE | 76.12 | 86.74 | 65.49 | 61.62 | 8.65 |

Tableau 2. Résultats comparatifs pour quelques relations verbales (f-mesures)

DICOVALENCE semble avoir des problèmes sur certains verbes très fréquents, en particulier sur le verbe *être*, comme le montre la très faible f-mesure pour la relation **ATB-SO**

(8.65% contre 66.46% pour L*efff,* cf. tableau 2) et comme également mis en évidence par la fouille d'erreurs.

## 5. Fouille d'erreurs

La fouille d'erreurs sur les verbes reprend les principes de fouilles d'erreurs présentés dans (Sagot & Villemonte de La Clergerie, 2006), tout en l'adaptant. En effet, dans ce travail, l'objectif était en première approximation d'identifier les formes dont la présence dans une phrase tend à rendre la phrase inanalysable[7]. Dans le présent article, cet objectif est adaptée afin d'identifier les entrées d'un lexique **hyp** qui semblent dégrader les performances de FRMG comparativement à un lexique de référence **ref**, ici le L*efff* : l'idée est de trouver les formes, et plus précisément les formes verbales, dont la présence dans une phrase *analysable avec le lexique de référence* **ref** tend à rendre cette phrase *inanalysable avec le lexique* **hyp**.

Le corpus EASy est trop petit pour obtenir des statistiques suffisantes pour un diagnostic complet des ressources. Mais l'algorithme ne nécessitant pas de disposer d'un corpus de référence (seule l'analysabilité, c'est-à-dire la couverture, étant exploitée), nous avons pu ajouter au corpus EASy environ 100K phrases avec AFP, Europar, Wikipedia et Wikisources, ce qui constitue le CPJ (Corpus Passage Jouet).

Nous avons regardé en détail les 15 premiers suspects dans *LGLex* afin de déterminer d'où proviennent les erreurs (entre parenthèses est indiqué le nombre de phrases contenant ce verbe qui n'ont pas pu être analysées, sachant que seule une phrase est donnée par verbe à titre d'exemple, mais qu'au total 212 phrases sont concernées pour cet échantillon) :

− Certaines entrées ne figurent pas dans les tables : c'est le cas de *réaffirmer* (28), de *réélire* (10), de la forme pronominale *se réimplanter* (5), mais également de *mixer* (7) dans la phrase *Mixé par Jimi Hazel , assisté de Bruce Calder , enregistré chez Jimi à l' « Electric Lady Studios » à New York* puisque cette entrée est codée dans la table 36S avec un sens différent (*Max mixe les carottes (et+avec) les navets dans un mixeur*).

− Certaines entrées figurent dans les tables mais ne sont pas codées (codage ~) : c'est le cas de *susciter* (41) qui n'est codée dans aucune des deux tables dans laquelle elle est présente (36DT et 38R), *recruter* (14) qui figure dans la table 38R sans être codée (ce qui implique qu'à part la construction de base N0 V N1 Prép N2, aucune autre construction n'est codée), *délocaliser* (9) qui figure dans la table 38L sans être codée (sa construction de base est N0 V N1 Loc N2 source Loc N3 destination, les effacements de certains compléments pouvant être codés dans la table), et *zapper* (4) qui figure dans la table 35L sans être codée (sa construction de base est N0 V Loc N1 source Loc N2 destination, elle ne peut donc pas être reconnue dans la phrase *Elle a également " déploré " la mémoire de " plus en plus sélective " de la jeune femme , " qui zappe les détails qui font désordre "*).

− D'autres sont codées dans les tables mais avec des compléments obligatoires qui ne sont pas présents dans les phrases du corpus :
  − *kidnapper* (12) et *revendre* (5) dans des phrases sans deuxième complément, telles que *Les deux Italiens ont été kidnappés le 18 décembre* et dans *Charles mangeait l'avoine des chevaux , doublant les fournitures , revendant par une porte de derrière ce qui entrait par la grande porte* : ces deux entrées sont codées dans la table 36DT, elles acceptent comme construction de base N0 V N1 Prép N2, sans effacement possible du deuxième complément introduit par la préposition *à* (c'est le cas de toutes les entrées de cette table) ;

---
[7]Une forme suspecte doit aussi avoir tendance à apparaître seule dans de telles phrases ou en cooccurrence avec des formes qui ne sont pas (trop) suspectes à l'échelle du corpus entier. On trouvera le détail du modèle sous-jacent dans (Sagot & Villemonte de La Clergerie, 2006) ; il s'exprime sous forme d'une paire d'équations mutuellement récursives reflétant le niveau local des phrases et le niveau global du corpus. L'algorithme de résolution est un algorithme de point fixe.

- *écrouer* (5) dans la phrase *Le lycéen de 18 ans soupçonné d'avoir poignardé vendredi un camarade , Hakim , dans leur lycée du Kremlin-Bicêtre \( Val-de-Marne \) , a été mis en examen et écroué hier , alors que lycées et collèges sont invités à observer une minute de silence aujourd'hui à la mémoire de la victime* : cette entrée est codée dans la table 38LHD avec la construction N0 V N1 Loc N2 destination ;
- *réprouver* (11) dans la phrase *Dieu ne réprouve donc personne* : cette entrée est codée dans la table 12 avec la construction N0 V N1 de N2.

− Enfin, certains cas spécifiques : *rediriger* (50) dans des phrases erronées, telles que *deux cent cinquante-troisredirige ici*, *consoler* (6) dans des phrases avec pronominalisation de l'objet, telles que *Elle essayait de le consoler* (l'entrée est codée dans les tables 13 et 32R1 avec la construction N0 V N1 acceptée, mais sans qu'aucune ne code la possibilité de pronominaliser le premier complément), et *camper* (5) dans *Les troupes campent entre Harlem et Nimègue* (l'entrée devrait être reconnue car elle accepte la construction N1 V).

Si l'on se penche à présent sur les 5 premiers verbes les plus suspects dans les 9 phrases suivantes qui n'ont pas été analysées avec le nouveau Le*fff*, l'origine de l'erreur est plus difficile à déterminer et ne semble pas toujours liée au nouveau Le*fff*. Il se peut, dans certains cas, que le succès de l'analyse avec le Le*fff* standard soit plutôt le résultat d'une surgénération :

− *tomber* employé dans une phrase complexe avec partie du corps : *cette lourde pensée lui tombe sur le coeur* ;
− *tomber* et *dominer* employés dans une construction avec *laisser*, qui semble mal gérée : *Reprit M Levrault en se laissant tomber dans un fauteuil* ; *moi je dis qu' il faut pas laisser JS tomber comme ça* ; *Pour quelles raisons, en temps de guerre, un très grand nombre de personnes, habituellement paisibles et inoffensives, se laissent-elles dominer par la haine* ;
− *cuisiner*, *approcher* et *orner*, employés dans des phrases pour laquelle l'échec n'est pas imputable aux entrées verbales : *il sait pas cuisiner* ; *Comme il approchait du château de ses pères* ; *elle a voulu à tout prix, orner sa vieille ville* ; *elle l' a ornée, sans proportion avec ses destinées et son avenir* ;
− *parer*, à qui il manque la redistribution passive : *Partout, sur la route, les fenêtres vous regardent parées de fleurs et de verdure*.

## 6. Conclusion

Convertir un lexique au format Le*fff* permet de l'utiliser à peu près immédiatement avec FRMG. Pour un lexique de bonne qualité comme DICOVALENCE ou les tables du Lexique-Grammaire, les résultats obtenus sont bons. En effet, une f-mesures pour les relations au dessus de 60% est meilleure que les résultats de FRMG avec le Le*fff* lors de la campagne de 2007 (59,65% de f-mesure pour 56% de couverture). Néanmoins, les derniers points de f-mesure découle d'une adaptation plus fine entre la grammaire et le lexique, et de la recherche des erreurs ou incomplétudes lexicales. Il est en effet normal que tout lexique possède des entrées erronées qu'il est difficile de trouver.

Les techniques de fouille d'erreurs permettant de comparer (dans un sens ou l'autre) les verbes de deux lexiques sont un moyen de repérer plus rapidement ces entrées.

À terme, ce travail doit aussi renforcer la fusion de diverses ressources lexicales en une seule ressource de très grande qualité. Néanmoins, les choix linguistiques derrière chaque ressource ont un impact : ainsi, le Le*fff* standard (version 3) fournit des entrées verbales plutôt factorisées (peu de distinctions sémantiques, cadres de sous-catégorisation factorisés) alors que *LGLex* liste de nombreuses entrées par verbes correspondant à divers sens et associées à des cadres de sous-catégorisation plus simple mais se chevauchant.